\documentclass[10pt,twocolumn,letterpaper]{article}

\usepackage{iccv}
\usepackage{times}
\usepackage{epsfig}
\usepackage{graphicx}
\usepackage{amsmath}
\usepackage{amssymb}

\usepackage{comment}

\usepackage[pagebackref=true,breaklinks=true,letterpaper=true,colorlinks,bookmarks=false]{hyperref}

\iccvfinalcopy 

\def\iccvPaperID{315} 

\ificcvfinal\fi
\begin{document}

\title{Improving Image Classification with Location Context}

\author{Kevin Tang$^{1}$, \;  Manohar Paluri$^2$, \; Li Fei-Fei$^1$, \; Rob Fergus$^2$, \; Lubomir Bourdev$^2$\\
$^1$Computer Science Department, Stanford University
~~~~~
$^2$Facebook AI Research\\
{\tt\small \{kdtang,feifeili\}@cs.stanford.edu}
~~~~~
{\tt\small \{mano,robfergus,lubomir\}@fb.com} \\
\url{https://sites.google.com/site/locationcontext/}
}

\maketitle

\begin{abstract}
With the widespread availability of cellphones and cameras that have GPS capabilities, it is common for images being uploaded to the Internet today to have GPS coordinates associated with them. In addition to research that tries to predict GPS coordinates from visual features, this also opens up the door to problems that are conditioned on the availability of GPS coordinates. In this work, we tackle the problem of performing image classification with location context, in which we are given the GPS coordinates for images in both the train and test phases. We explore different ways of encoding and extracting features from the GPS coordinates, and show how to naturally incorporate these features into a Convolutional Neural Network (CNN), the current state-of-the-art for most image classification and recognition problems. We also show how it is possible to simultaneously learn the optimal pooling radii for a subset of our features within the CNN framework. To evaluate our model and to help promote research in this area, we identify a set of location-sensitive concepts and annotate a subset of the Yahoo Flickr Creative Commons 100M dataset that has GPS coordinates with these concepts, which we make publicly available. By leveraging location context, we are able to achieve almost a 7\% gain in mean average precision.
\end{abstract}

\section{Introduction}

\begin{figure}[t]
\begin{center}
\includegraphics[width=1\linewidth]{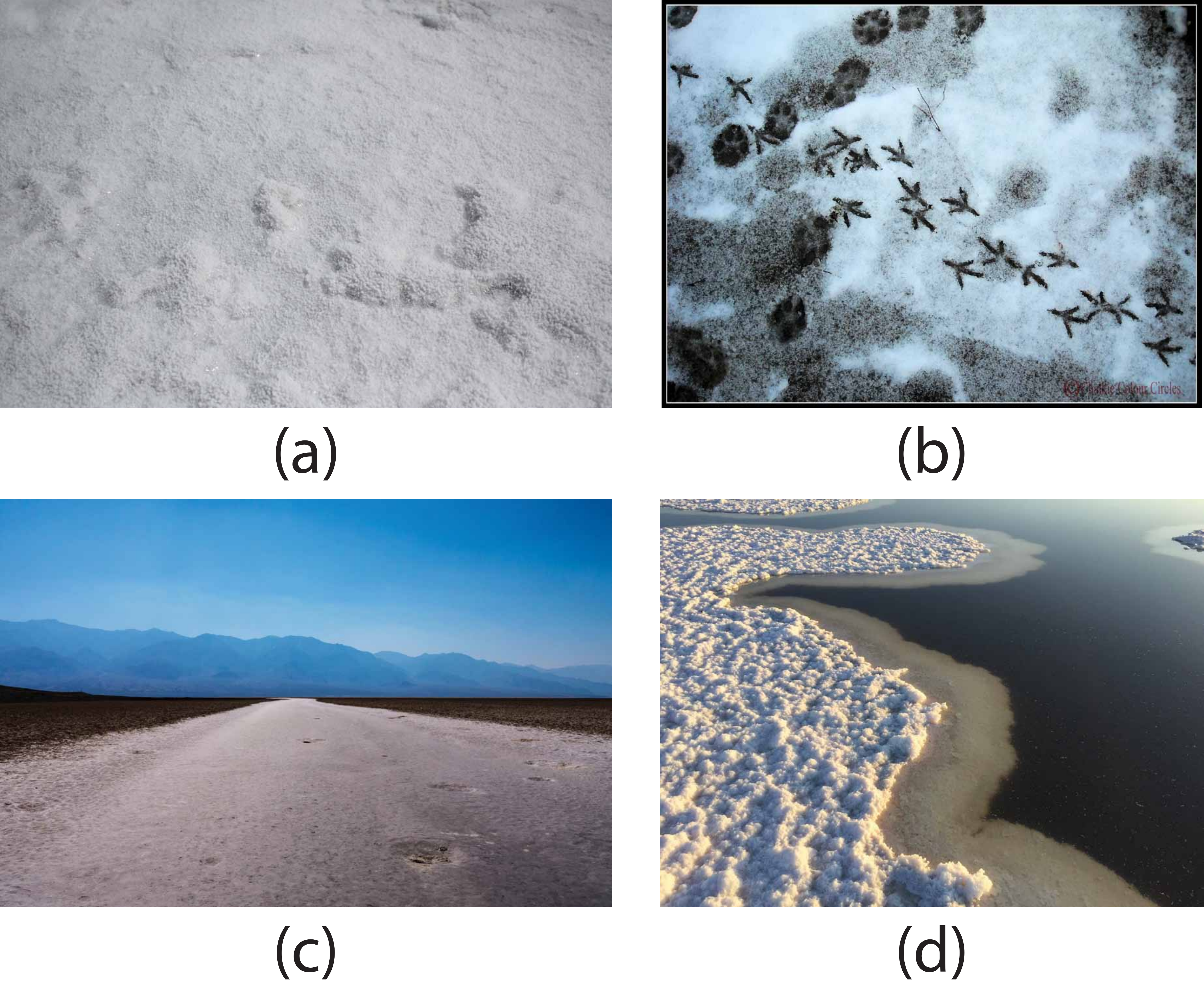}
\end{center}
   \caption{Which of these are images of snow? Just by looking at the images, it may be difficult to tell. However, what if we knew that (a) was taken at the Bonneville Salt Flats in Utah, (b) was taken in New Hamsphire, (c) was taken in Death Valley, California and (d) was taken near Palo Alto, California? Image credits given in supplementary material. \vspace{-4mm}}
\label{fig:pullfig}
\end{figure}

As Figure~\ref{fig:pullfig} shows, it is sometimes hard even for humans to recognize the content of photos without context. Just by looking at the photos we can conclude that all these examples can reasonably be of snow. Consider, however, that (a) was taken at the Bonneville Salt Flats in Utah, (c) and (d) were taken in Death Valley and Palo Alto, respectively, both of which are areas in California that never see snow, and (b) was taken in New Hampshire, where snow storms are common. With this information in hand, it is much easier to correctly deduce that (b) is the only image that actually contains snow.

Motivated by this observation, we tackle the problem of image classification with location context. In particular, we are interested in classifying consumer images with concepts that commonly occur on the Internet, ranging from objects to scenes to specific landmarks, as these are the things that people often take pictures of, and the Internet is the largest source of geotagged images. Building on the CNN architecture introduced in~\cite{alexnet}, the basis for most state-of-the-art image classification and recognition results, we address how to represent and incorporate location features into the network architecture. This is not an easy problem, as we have found that naive approaches such as concatenating the GPS coordinates into the classifier, or leveraging nearby images as a Bayesian prior result in almost no gain in performance. However, knowing the GPS coordinates allows us to utilize geographic datasets and surveys that have been collected by various institutions and agencies. We can also leverage the large amount of data on the Internet tagged with GPS coordinates in a data-driven fashion.

In summary, the contributions in this paper can be organized into three parts.

\vspace{-3mm}

\paragraph{Constructing effective location features from GPS coordinates.}
We propose 5 different types of features that extend upon the latitude and longitude coordinates that are given to us, and perform a comprehensive evaluation of the effectiveness of each feature.

\vspace{-3mm}

\paragraph{Network architectures for incorporating location features.}
We show how to incorporate these additional features into a CNN~\cite{alexnet}. This allows us to learn the visual features along with the interactions between the different feature types in a joint framework. In addition, we also show how we can simultaneously learn the parameters required for constructing a subset of our features in the same framework, giving us improved performance and a better understanding of what the network is learning.

\vspace{-3mm}

\paragraph{YFCC100M-GEO100 dataset.}
We introduce annotations for a set of location sensitive concepts on a subset of the Yahoo Flickr Creative Commons 100M (YFCC100M) dataset~\cite{yfcc100m}, which we denote as the YFCC100M-GEO100 dataset, and make our annotations publicly available. This dataset consists of 88,986 images over 100 classes, and allows us to evaluate our models at scale.

\section{Related Work}

There is a large body of work that focuses on the problem of image geolocation, such as geolocating static cameras~\cite{Jacobs07}, city-scale location recognition~\cite{Schindler07}, im2gps~\cite{Hays08,Zamir10}, place recognition~\cite{Gronat13,Torii13}, landmark recognition~\cite{Bergamo13,Chen11,Raguram11,Zheng09}, geolocation leveraging geometry information~\cite{Bergamo13,Irschara09,Li10,Sattler12}, and geolocation with graph-based representations~\cite{Cao13}. More recent works have also tried to supplement images with corresponding data~\cite{Baatz12,Bansal14,Lee15,Li14,Lin13}, such as digital elevation maps and land cover survey data, which we draw inspiration from in constructing our features. In contrast to these works, we assume we are given GPS coordinates, and use this information to help improve image classification performance.

In addition, several works also explore other aspects of images and location information, such as 3D cars with locations~\cite{Matzen13}, organizing geotagged photos~\cite{Crandall09}, structure from motion on Internet photos~\cite{Snavely08}, recognizing city identity~\cite{Zhou14}, looking beyond the visible scene~\cite{Khosla14}, discovering representative geographic visual elements~\cite{Doersch12,Lee13}, predicting land cover from images~\cite{Leung10}, and annotation enhancement using canonical correlation analysis~\cite{Cao09}.

Most similar are works that leverage location information for recognition tasks~\cite{Amlacher09,Ardeshir14,Berg14,Yu08}. The work of~\cite{Amlacher09} tackles object recognition with geo-services on mobile devices in small urban environments. The work of~\cite{Ardeshir14} uses available Geographic Information System (GIS) databases by projecting exact location information of traffic signs, traffic signals, trash cans, fire hydrants, and street lights onto images as a prior. The work of~\cite{Berg14} uses bird sightings to estimate a spatio-temporal prior distribution to help improve fine-grained categorization performance. The work of~\cite{Yu08} leverages season and location context in a probabilistic framework to help improve region recognition in images. Our work differs in that we are interested in recognizing a wide range of concepts present on the Internet beyond birds~\cite{Berg14}, small sets of specific urban objects~\cite{Amlacher09,Ardeshir14} or generic region types~\cite{Yu08}, and constructing features that are not specific to a particular class or source of GIS information. In addition, we exhaustively evaluate ways of incorporating these features into a CNN, and we propose a way to parameterize the geo-features and extend the back-propagation algorithm to allow the net to learn the most discriminative geo-feature parameters. We also introduce a large-scale geotagged dataset collected from real-world images to train our models and effectively evaluate performance. 

Also closely related are the numerous works on context, which have shown to be helpful for various tasks in computer vision~\cite{Divvala09,Torralba03ijcv}. We leverage contextual information by considering the GPS coordinates of our images and extracting complementary location features.

\begin{figure*}
\begin{center}
\includegraphics[width=1\linewidth]{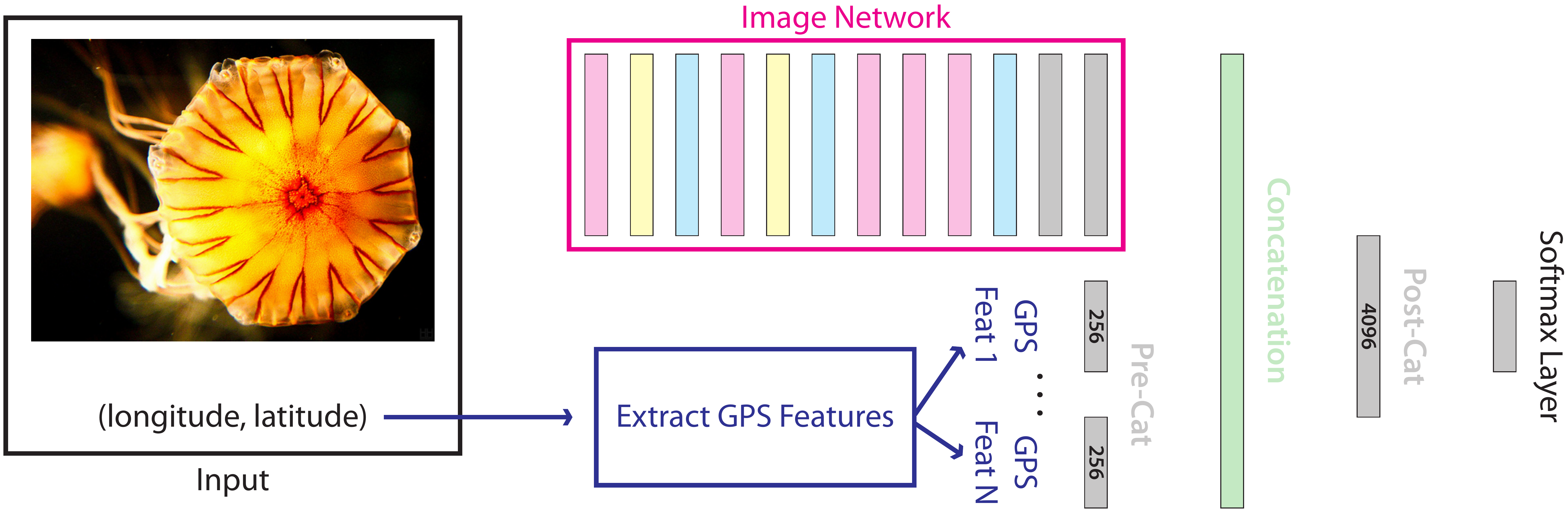}
\end{center}
   \caption{Our CNN architecture. The pink rectangles denote convolutional layers, the yellow rectangles denote normalization layers, the blue rectangles denote pooling layers, the grey rectangles denote fully connected layers, and the green rectangles denote concatenation layers. The final fully connected layer is the softmax layer. Our model is given as input an image and its associated longitude and latitude coordinates. The image network denoted by the magenta box is the network architecture introduced in~\cite{alexnet}. \vspace{-4mm}}
\label{fig:model}
\end{figure*}

\section{Our Approach}

Similar to standard image classification problems, we are given a set of $n$ training images $\{I_1, I_2, \ldots, I_n\}$ with associated class labels $\{y_1, y_2, \ldots, y_n\}$, where $y \in \mathcal{C}$ is the set of classes we are trying to predict. In addition to the images, we are also given the GPS coordinates for each image $\{(long_1, lat_1), (long_2, lat_2), \ldots, (long_n, lat_n)\}$, where $long_i$ is the longitude and $lat_i$ is the latitude for image $i$. Note that the GPS coordinates are given in both the training and testing phase, and our goal is to predict the class labels given both the image and the GPS coordinates. In this paper, we focus on images taken within the contiguous United States, but the majority of our features can be trivially extended to encompass the entire world.

\subsection{Neural network architecture}
We build on the CNN model introduced in~\cite{alexnet}, as this model and extensions to it are commonly used benchmarks in image classification and recognition~\cite{Girshick14,imagenet,googlenet}. For more details on the network architecture, we refer the reader to~\cite{alexnet}. To incorporate location features into the network, we add a layer to concatenate the different feature types before the softmax layer, as shown in Figure~\ref{fig:model}. This makes intuitive sense, as the lower layers of the CNN model are aimed at learning effective image filters and features, and we are interested in incorporating our features later on at a higher semantic level. In addition, we also experiment with adding additional depth using fully connected layers before and after the concatenation layer, denoted by the pre-cat and post-cat layers in Figure~\ref{fig:model}, and perform comprehensive experiments detailed later in the paper.

With this architecture, we turn to the problem of extracting location features. For each image $i$, we construct a set of features that effectively represent contextual information about the location specified by the GPS coordinates $(long_i, lat_i)$. To do this, we utilize the wealth of geographic datasets and surveys collected by various agencies that document a large variety of statistics about each location, ranging from surveys on age and education, to geographic features such as elevation and precipitation. We also utilize the large amounts of geotagged data available on the Internet such as images and textual posts.

\subsection{GPS encoding feature}
The actual GPS coordinates are very fine location indicators, making it difficult for the classifier to effectively use. To make better use of the coordinates, we grid the contiguous United States into a rectangular grid with a latitude to longitude ratio of $\frac{1}{2}$, and construct an indicator vector for each image $i$ that indicates which grid cell the GPS coordinate $(long_i, lat_i)$ falls into, resulting in a feature vector with dimension equal to the total number of cells in the grid. The aspect ratio is chosen so that each grid cell is roughly a square. We used rectangular grids up to $100$x$200$, resulting in $25$x$25$km square cells, limited by the computational time and memory for even larger grids.

\begin{figure}[t]
\begin{center}
\includegraphics[width=0.95\linewidth]{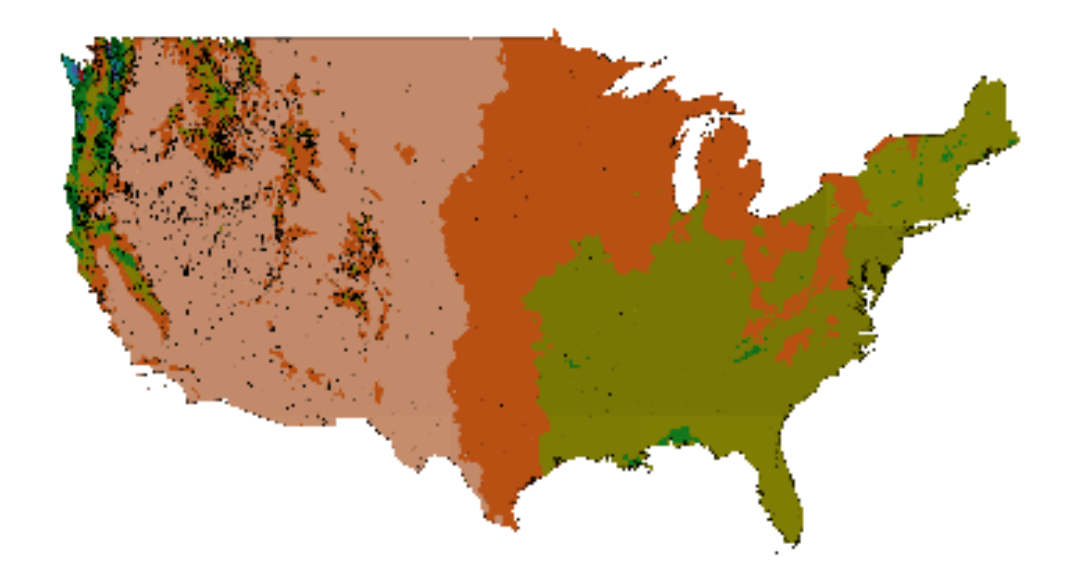}
\end{center}
   \caption{Example geographic map of precipitation in the United States~\cite{googlemaps}, with darker colors roughly indicating larger values of average precipitation. Regions with more rainfall may give rise to images that more commonly contain objects such as umbrellas. \vspace{-4mm}}
\label{fig:mapfeat}
\end{figure}

\subsection{Geographic map feature}
There exist many different types of geographic maps and datasets that provide detailed information about each GPS coordinate in the form of a colored map, with different colors representing different geographic features. In particular, Google Maps~\cite{googlemaps} is one of many online sites that stores a large set of such maps, with an example shown in Figure~\ref{fig:mapfeat}. We use 10 different types of maps from Google Maps: average vegetation, congressional district, ecoregions, elevation, hazardous waste, land cover, precipitation, solar resource, total energy, and wind resource. Since each map uses different colors to represent the value of a feature at a particular location, for each image $i$ we take the normalized pixel color values in a 17x17 patch around the GPS coordinate $(long_i, lat_i)$ for each map type, and concatenate these to form a 8670 dimensional feature. Intuitively, map features such as precipitation may tell us how likely it is to see an umbrella in a picture, while indicators such as elevation may tell us how likely it is for us to see snow.

\subsection{ACS feature}
Given the GPS coordinate $(long_i, lat_i)$ for image $i$, we can perform reverse geocoding to obtain the corresponding zip code. This allows us to tap into the rich source of geographic surveys organized by zip code, collected by agencies like the United States government. We use the American Community Survey (ACS)~\cite{acs}, an ongoing survey that provides yearly data with statistics on age, sex, race, family/relationships, income and benefits, health insurance, education, veteran status, disabilities, work status, and living conditions, all organized by zip code and pooled over a 5 year period. We treat each statistic as a feature, and collect them into a vector resulting in a 21,038 dimensional feature. Intuitively, statistics such as age may tell us how likely it is to see toys in a picture, while statistics such as income may tell us how likely it is for us to see expensive cars.

\begin{figure}[t]
\begin{center}
\includegraphics[width=0.95\linewidth]{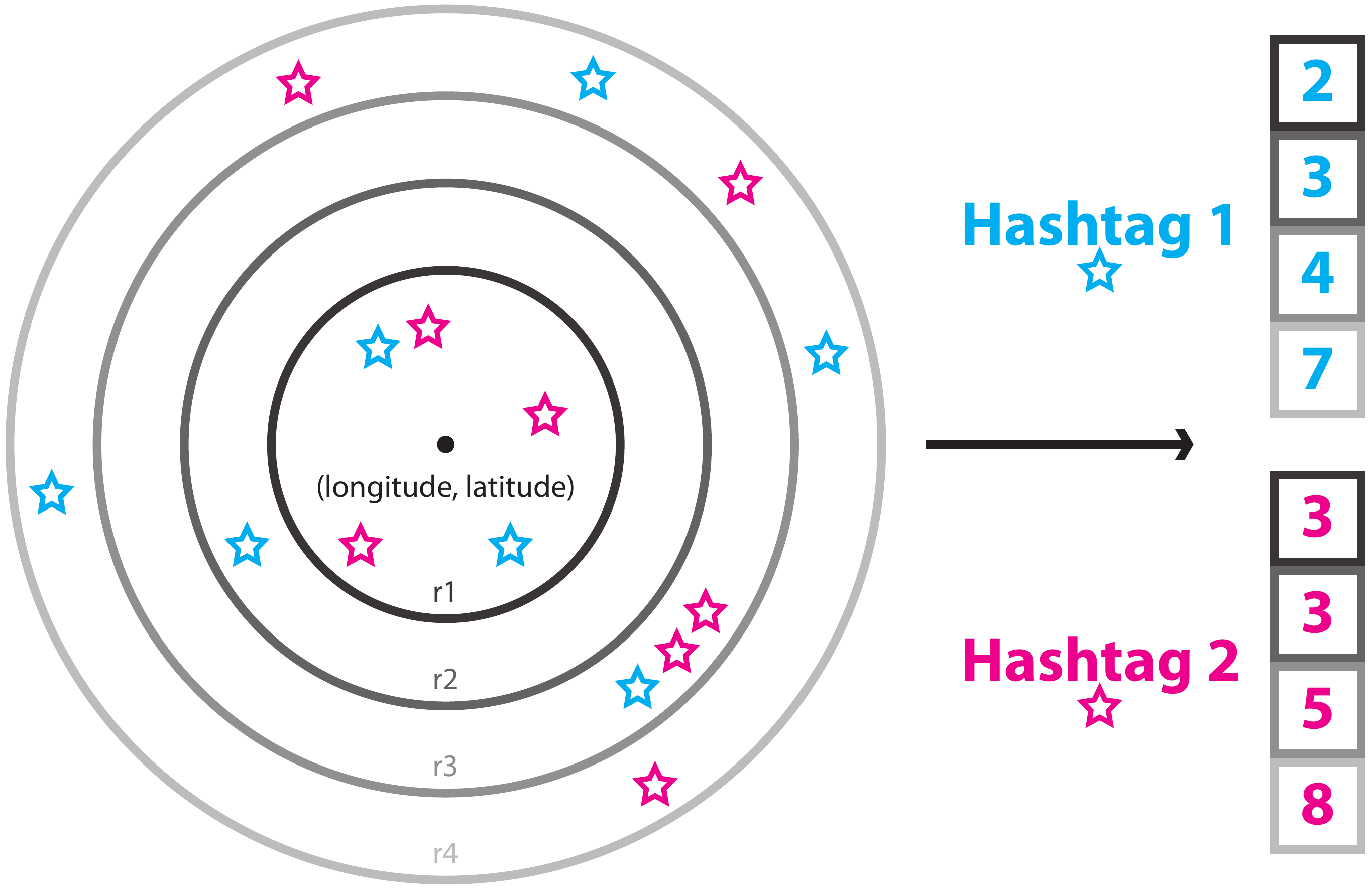}
\end{center}
   \caption{To build the hashtag context features, we look at the distribution of hashtags around each GPS coordinate by finding Instagram images tagged with relevant hashtags and GPS coordinates. For each hashtag, we pool over circles of different radii, counting the number of times each hashtag (blue/magenta stars) appears within a particular radius. \vspace{-4mm}}
\label{fig:gridfeat}
\end{figure}

\subsection{Hashtag context feature}
The aforementioned geographic map and ACS features are based on map and survey data collected about a particular location from various agencies. However, a large source of data lies directly on the Internet, where millions of images are uploaded daily, many of which tagged with GPS coordinates. We propose a set of data-driven features that are able to make use of the images on Instagram~\cite{instagram}.

Intuitively, for each image $i$ associated with GPS coordinate $(long_i, lat_i)$, our goal is to capture the distribution of hashtags in the vicinity. Hashtags that commonly occur near image $i$ can help indicate the types of things that occur in the real-world context of image $i$, giving us contextual information about what is in the image. We start by defining a set of hashtags $\mathcal{H}$ that we are interested in. For a particular hashtag $h \in \mathcal{H}$, we obtain images from Instagram with GPS coordinates and matching hashtag. Then, we define a set of radii $\mathcal{R}$, and for each $r \in \mathcal{R}$, we pool over a circle of radius $r$ around $(long_i, lat_i)$ and count the number of images tagged with hashtag $h$ that fall into the radius. As shown in Figure~\ref{fig:gridfeat}, this is done for each of the radii in $\mathcal{R}$ and each of the hashtags in $\mathcal{H}$, resulting in a set of $|\mathcal{H}|$x$|\mathcal{R}|$ counts.

To build features from these counts, we perform two types of normalization for the $|\mathcal{H}|$ counts in each radius $r \in \mathcal{R}$. The first is normalization across hashtags, where we normalize each count by the sum of counts for all hashtags within $r$. This normalization gives us an idea of the relative frequency of a particular hashtag in relation to the other hashtags that appear in the area, and normalizes for the density of photos in the area. The second is normalization within hashtag, where we normalize each count by the total number of images we obtained from Instagram with the particular hashtag. This normalization gives us an idea of the relative frequency of a particular concept in relation to how often this concept appears in the entire United States. We perform both types of normalization and concatenate the feature vectors together to form the final feature vector, resulting in a $2$x$|\mathcal{H}|$x$|\mathcal{R}|$ dimensional feature.

In our experiments, we set $\mathcal{C} = \mathcal{H}$, using the set of classes as the set of hashtags for simplicity, and set $\mathcal{R} = \{1000,2000, \ldots, 10000\}$. To save computation time, we quantize all the GPS coordinates into a $25000$x$50000$ grid, which results in approximately square grid cells each covering a $100$x$100$ meter area.

\subsection{Visual context feature}
The visual context feature is similar to the hashtag context feature, except in this case we would like to take advantage of the visual signal around our GPS coordinate $(long_i, lat_i)$, and not just the hastags that have been tagged. To do this, we retrieve images from various online social websites with GPS coordinates, and for each image run a CNN with similar architecture to~\cite{alexnet} to generate probabilities for 594 of the common types of concepts that appear on the Internet, such as ``clothes", ``girl", and ``coffee". The full list of 594 concepts is given in the supplementary material.

Similar to the hashtag context feature, we pool the probabilities for each radius $r \in \mathcal{R}$ around $(long_i, lat_i)$ by summing the probabilities of all the images that fall into the radius, individually for each concept, resulting in a set of $594$x$|\mathcal{R}|$ probabilities. Then, we perform the same two types of normalization and concatenate to form the final feature vector, resulting in a $2$x$594$x$|\mathcal{R}|$ dimensional feature. We use the same set of radii $\mathcal{R}$ and GPS grid quantization as the hashtag context feature.

\section{Learning the Optimal Pooling Radius}

In the previous section we introduced the hashtag context and visual context features. For both of these features, we explained how to build features from the aggregated hashtag counts and concept probabilities by concatenating together normalized histograms pooled over a set of radii $\mathcal{R}$. However, we don't expect that all radii are informative. For example, for hashtags or concepts that are rare, even being a few kilometers away may be an important indicator. Similarly, certain hashtags that are common may require being extremely close to truly pinpoint the location.

\begin{figure*}
\begin{center}
\includegraphics[width=1\linewidth,height=2.5cm]{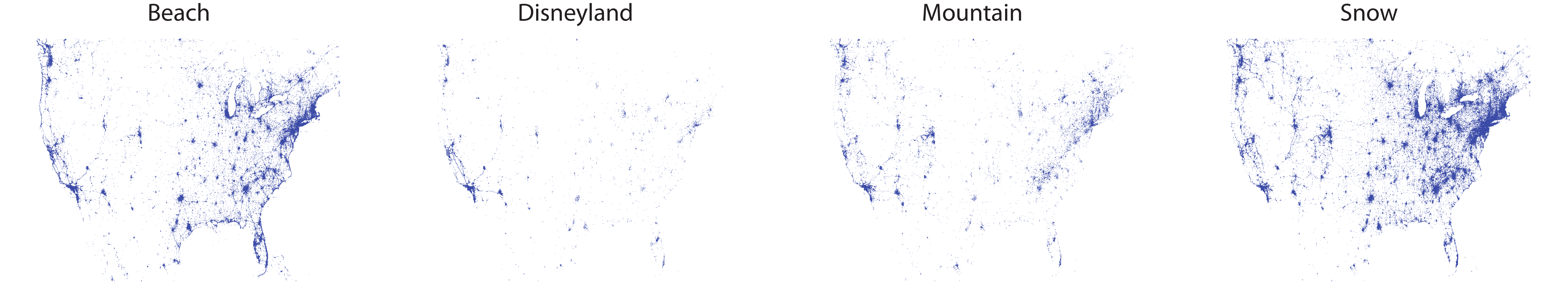}
\end{center}
   \caption{Instagram hashtag distributions for various classes in the contiguous United States. Although we can see interesting patterns such as beach hashtags near coasts and the outline of the Appalachian Mountains in the mountain hashtags, there is a great deal of noise. \vspace{-4mm}}
\label{fig:kl}
\end{figure*}

\vspace{-3mm}

\paragraph{Radius learning layer.}
To address this, we show how to construct a layer in the CNN that automatically learns the optimal radius used for pooling, which we denote as the radius learning layer. Learning the optimal radius is potentially useful in many ways. First, by focusing on the important radii with informative features, we can avoid overfitting. Second, we can visualize the radii that we have learned, providing insight into what the CNN is learning.

We start by considering the radius for a single hashtag/concept $h$, and fit a function $H_{(long_i, lat_i),h}(\rho)$ over the histogram that returns the value of the histogram feature for hashtag $h$ and radius $\rho$ at location given by $(long_i, lat_i)$. There are several ways to fit such a function, but for simplicity we use the histogram values computed over $\mathcal{R}$ from the previous section and fit a piece-wise linear approximation to the values. We do this for all the hashtags, concepts, and both types of normalization schemes to obtain a set of $2 \cdot (\mathcal{H} + 594)$ histogram functions for each training image $i$.

The outputs of these histogram functions are treated as input features to the CNN in place of the concatenated histograms, with a radius parameter $\rho_h$ for each hashtag/concept that selects the value of the function to treat as input to the neural network. When computing the gradient for backpropagation, we backpropogate the gradient of the error $E$ into the gradient of the histogram function $H$:

\begin{align}
\frac{\partial E}{\partial \rho_h} = \frac{\partial E}{\partial H_{(long_i, lat_i),h}(\rho_h)} \frac{\partial H_{(long_i, lat_i),h}(\rho_h)}{\partial \rho_h}
\label{eqn:radius_bp}
\end{align}

The first term in the RHS is the error derivative propagated to the radius learning layer from the network architecture above it, and the second term is the derivative at $\rho_h$ of the histogram function $H_{(long_i, lat_i), h}$. Since we use a piece-wise linear approximation to fit the histogram function, the second term is easily computed by taking the slope between the two nearest points in $\mathcal{R}$. Although we could fit more complicated functions, we found the linear approximation to be fast and sufficient, as we aggregate gradients over all the training examples. Since hashtags/concepts may have multiple radii and weightings between the radii that are informative, we replicate the radius learning layer multiple times for each histogram function.

\section{Dataset}

To evaluate our method, we use the recently released Yahoo Flickr Creative Commons 100M (YFCC100M) dataset~\cite{yfcc100m}, which consists of 100 million Creative Commons copyright licensed images from Flickr. Of the 100 million images, approximately 49 million are geotagged with GPS coordinates, which makes this dataset particularly suitable for evaluating our task because of its unprecedented scale. Also provided with the images are tags for the images produced by users on Flickr, which we use as a first step to identify images that contain a particular class. However, because the tags are very noisy, we must manually verify and discard images that do not actually contain the classes we were interested in. As mentioned before, we focus only on images geotagged within the contiguous United States.

\vspace{-3mm}

\paragraph{Selecting location-sensitive classes.}
One of the problems we have to deal with is selecting classes that are likely to be location-sensitive, and will benefit from our location context features. This is important because there are certainly classes that are not, and adding these additional features may just cause the classifier to overfit. Practically speaking, we also need a way of limiting the number of classes to a manageable number we can annotate.

To address this issue, we use a simple data-driven method for selecting classes. Using a large set of images from Instagram, we estimate the discrete geospatial distribution $P$ of all images by first gridding the contiguous United States into a fine grid, and then counting the number of images that fall into each grid cell and normalizing to create a valid probability distribution. Then, we obtain a large list of classes through commonly occuring Instagram hashtags, and for each class $c$ we estimate the geospatial distribution $Q_{c}$ of images tagged with $c$ in a similar manner. With these two distributions, we compare their similarity with the Kullback-Leibler (KL) divergence:

\begin{align}
D_{KL}( P || Q_{c}) = \sum_i P(i) \ln \frac{P(i)}{Q_{c}(i)}
\label{eqn:kl}
\end{align}

Intuitively, we would like to find classes that do not exhibit a geospatial distribution similar to the distribution of all images, as this would suggest that they have some location-sensitive properties. The KL divergence does this by giving us a measure of the difference between the two probability distributions, and we select the top 100 classes with the highest KL divergence. In practice, given a new class $c$, we can simply compute $D_{KL}(P || Q_{c})$ and threshold to determine whether or not the class will benefit from our additional location features. Examples of the geospatial distributions are shown in Figure~\ref{fig:kl}.

\begin{figure}[t]
\begin{center}
\includegraphics[width=1\linewidth, height=5cm]{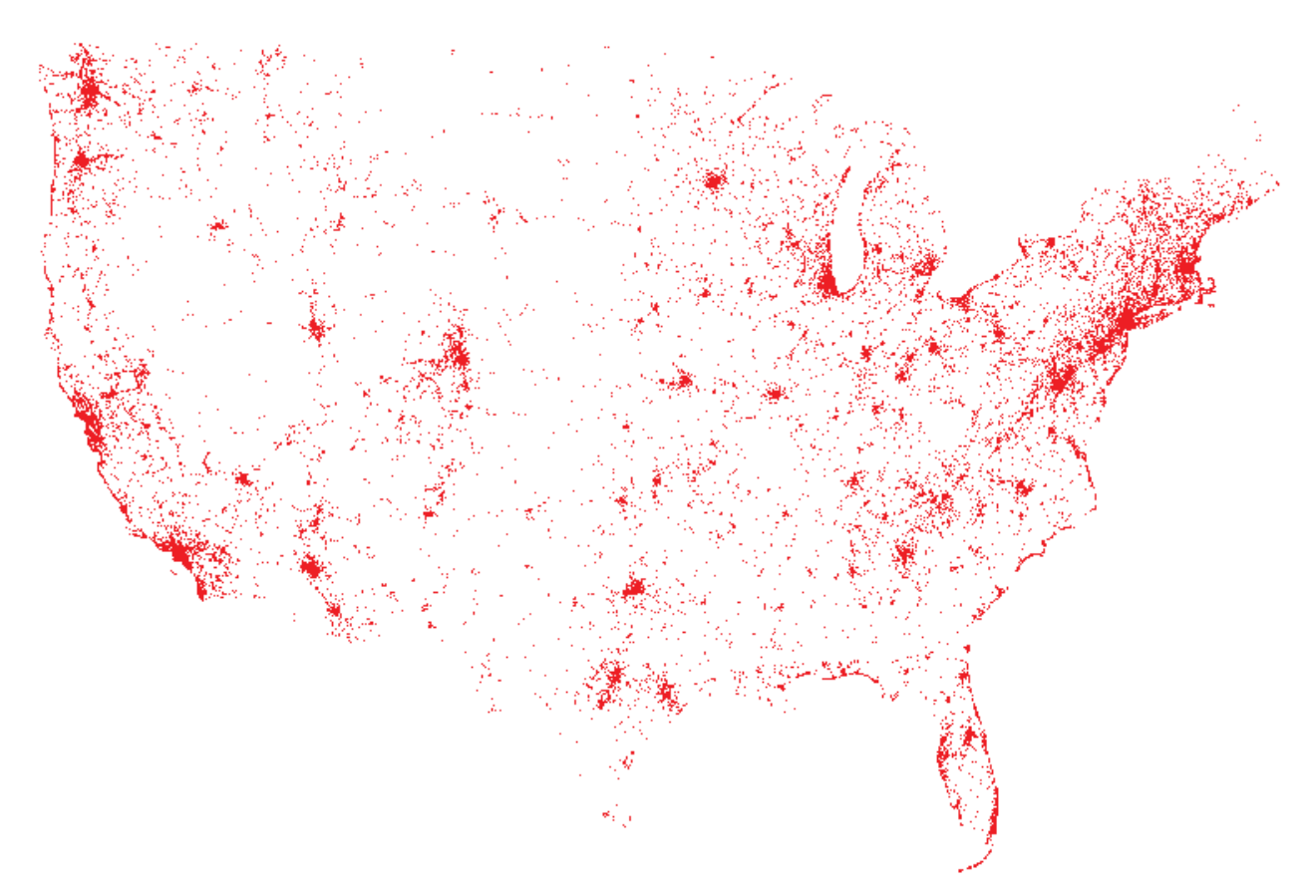}
\end{center}
   \caption{The geographic distribution of the 88,986 images in the YFCC100M-GEO100 dataset that we introduce. \vspace{-4mm}}
\label{fig:yfcc100m}
\end{figure}

\vspace{-3mm}

\paragraph{YFCC100M-GEO100 dataset.}
Using the top 100 classes selected with the highest KL divergence, we manually verified and annotated a large set of the YFCC100M images that were noisily tagged with these classes by Flickr users. This resulted in a dataset of 88,986 images, with at least 100 images per class, which we denote as the YFCC100M-GEO100 dataset and will make publicly available. The classes we selected range from objects to places to scenes, with examples such as `autumn', `beach', and `whale' that illustrate the diversity of classes we are trying to classify. Figure~\ref{fig:yfcc100m} visualizes the distribution of GPS coordinates for the images in the dataset. The full list of classes is given in the supplementary material.

\section{Results}

We randomly divide the YFCC100M-GEO100 dataset into an 80\% training set and 20\% test set. We further leave out a small portion of the training set as a validation set for parameter tuning in our models.

\vspace{-3mm}

\paragraph{Implementation details.}
Following~\cite{alexnet}, we train our models using stochastic gradient descent with momentum of 0.9 and a 0.005 weight decay. We use a learning rate of $0.1$, and run approximately 30 passes through our data, decreasing the learning rate by $0.1$ every 10 passes. We use a $0.5$ dropout ratio for all of our fully connected layers. Since our training data is relatively small, we initialize the parameters in the Image Network portion of the model (see Figure~\ref{fig:model}) by pre-training it on a large set of Instagram images, and then freezing the pre-trained parameters into our model. Note that we could further fine-tune these parameters as well, but chose not to for speed concerns.

\vspace{-3mm}

\paragraph{Performance metrics.}
To evaluate our models, we use three different performance metrics. In addition to the standard metric of mean average precision (AP), we also include results for normalized accuracy@1 and normalized accuracy@5, motivated by their use in recent papers~\cite{Karpathy14} as well as the ImageNet classification challenge~\cite{imagenet}. The normalized accuracy@k measure indicates the fraction of test samples that contained the ground truth label in the top k predictions, normalized per class to adjust for differences in the number of images per class.

\begin{table}
\begin{center}
\resizebox{1\linewidth}{!}{
\begin{tabular}{|c|c|c|c|}
\hline
Method & Mean AP & Acc@1 & Acc@5 \\
\hline \hline
Image only & 36.82\% & 39.45\% & 70.15\% \\
Image + GPS coordinates & 36.83\% & 39.47\% & 70.23\% \\
Image + GPS encoding 10x20 & 38.58\% & 41.48\% & 72.39\% \\
Image + GPS encoding 100x200 & \textbf{38.89\%} & \textbf{41.67\%} & \textbf{72.47\%} \\
\hline \hline
Image + Geographic map feature & 37.70\% & 40.28\% & 70.79\& \\
Image + ACS feature & \textbf{40.41\%} & \textbf{42.79\%} & \textbf{73.84\%} \\
Image + Hashtag context feature & 39.86\% & 42.27\% & 73.38\% \\
Image + Visual context feature & 38.81\% & 41.53\% & 72.31\% \\
\hline \hline
Image only (SVM) & 33.41\% & 36.56\% & 60.05\% \\
Image + All features (SVM) & 34.61\% & 38.06\% & 62.88\% \\
Image + All features $\chi^2$ kernel (SVM) & \textbf{35.12\%} & \textbf{38.57\%} & \textbf{63.74\%} \\
\hline \hline
Image + Flickr prior 10NN & 24.15\% & 25.36\% & 36.46\% \\
Image + Flickr prior 100NN & 33.38\% & 35.45\% & 60.62\% \\
Image + Flickr prior 1000NN & \textbf{36.30\%} & \textbf{37.86\%} & \textbf{68.57\%} \\
\hline \hline
Image + Instagram prior 1000km & 24.03\% & 22.70\% & 38.23\% \\
Image + Instagram prior 4000km & 31.96\% & 30.62\% & 58.69\% \\
Image + Instagram prior 8000km & \textbf{33.08\%} & \textbf{30.67\%} & \textbf{60.13\%} \\
\hline
\end{tabular}
}
\end{center}
\caption{Results comparing various baseline methods. For the CNN models we do not use pre-cat and post-cat layers. \vspace{-4mm}}
\label{table:feat}
\end{table}

\subsection{Baseline methods}

We evaluate the benefit of each proposed feature, shown in the top two sections of Table~\ref{table:feat}, without the pre-cat layer and post-cat layers as a baseline (see Figure~\ref{fig:model}). Not surprisingly, using the GPS coordinates does not yield any significant gain in performance, as they do not make sense in the context of a linear classifier. Using the GPS encoding features, we get much better performance, with a gain of around 2\% in all performance measures. We can also see that for each feature, we obtain performance gains from concatenating the features with the baseline image features, which shows they provide complementary information. In particular, the ACS feature yields the largest increase in performance, with almost a 4\% gain in all performance measures.

\vspace{-3mm}

\paragraph{Support vector machines.}
We perform experiments using Support Vector Machines (SVM) and kernelizing our features. We use kernel averaging to combine features, as it has been shown to perform on par with more complicated methods~\cite{Gehler09}. In the middle section of Table~\ref{table:feat}, we show results using a multi-class hinge loss SVM classifier and cross-validating the regularization parameter. For the naive combination, we use linear kernels for all features, and for the $\chi^2$ combination, we compute $\chi^2$ kernels for the histogram features (hashtag context, visual context), and use linear kernels for the rest due to dimensionality concerns. In general, we found the SVM to perform worse than the softmax classifier. Kernelizing the histogram features with $\chi^2$ kernels performs better than using just linear kernels, but still doesn't exceed the performance of the softmax.

\vspace{-3mm}

\paragraph{Bayesian priors.}
Following the approach used in~\cite{Berg14}, we also try incorporating location context as a Bayesian prior. Using Bayes' rule, the probability of predicting class $c$ given image $I_i$ and location $(long_i, lat_i)$ can be written as:

\begin{align}
P(c | I_i, long_i, lat_i) = \frac{P(I_i, long_i, lat_i | c) P(c)}{P(I_i, long_i, lat_i)}
\label{eqn:bayes1}
\end{align}
Assuming the image and location are conditionally independent given the class, further applying Bayes' rule and removing terms that do not depend on $c$, we obtain:
\begin{align}
P(c | I_i, long_i, lat_i) \propto \frac{P(c | I_i)}{P(c)} P(c | long_i, lat_i)
\label{eqn:bayes2}
\end{align}
In our experiments, we assume a uniform prior over the classes for $P(c)$. We tried two different approaches to computing the location prior $P(c | long_i, lat_i)$, with results given in the bottom two sections of Table~\ref{table:feat}. In the Flickr prior, for each test image we find the $k$-nearest-neighbors ($k$-NN) from the training set in GPS space and use their labels to estimate a distribution for the location prior. In the Instagram prior, for each of our test images we take the histogram computed in the hashtag context feature for a certain radius $r$, and use the normalized histogram as the distribution for the location prior. Although the overall results do not improve for either method, it's interesting to note that for some classes such as ``disneyland", results improve by more than 45\% mean AP for both types of priors. However, for the majority of the classes, the location prior hurts rather than helps, causing an overall decrease in performance.

\begin{table}
\begin{center}
\resizebox{1\linewidth}{!}{
\begin{tabular}{|c|c|c|c|}
\hline
Method & Mean AP & Acc@1 & Acc@5 \\
\hline \hline
Image only & 36.82\% & 39.45\% & 70.15\% \\
Image + All features with -/- & 37.97\% & 40.19\% & 70.67\% \\
Image + All features with 128/- & 42.22\% & 44.76\% & 75.74\%\\
Image + All features with 256/- & 42.34\% & \textbf{44.82\%} & \textbf{75.86\%} \\
Image + All features with 512/- & 42.20\% & 44.43\% & 75.53\% \\
Image + All features with 1024/- & 41.60\% & 43.98\% & 75.16\% \\
\hline \hline
Image + All features with 256/4096 & \textbf{43.28\%} & 43.74\% & 74.30\% \\
\hline
\end{tabular}
}
\end{center}
\caption{Results when concatenating all features and varying the pre-cat and post-cat layers. The X/Y notation refers to the dimensionality X of the pre-cat layers and Y of the post-cat layer, with - representing no pre-cat or post-cat layer. \vspace{-4mm}}
\label{table:joint}
\end{table}

\subsection{Architectures for feature combination}
We evaluate the various architectures for combining the features together, and evaluate the effect of varying levels of depth before and after the concatenation layer in the model. Results are shown in Table~\ref{table:joint}. The top section of the table shows results for adding additional depth in the pre-cat layer for each individual feature, and the bottom section shows the result with a 4096 dimensional post-cat layer. We make all comparisons to the ``Image only" model from Table~\ref{table:feat}, which we now refer to as the baseline image model.

\vspace{-3mm}

\paragraph{Pre-cat layer.}
From the results, we see that simply concatenating the features together does not result in a significant increase in performance, likely because the feature dimension is large, and the model is overfitting. Thus, we introduce the pre-cat layers to capture relationships within each feature type, and to serve as dimensionality reduction. Although they perform comparably, the 256 dimensional layer seems to strike the best balance between performance and the number of parameters to learn, obtaining almost a 6\% gain in performance across all performance measures. We also tried adding additional depth beyond a single layer, but found that this did not help significantly and drastically increased the number of parameters to learn.

\vspace{-3mm}

\paragraph{Post-cat layer.}
We also perform experiments with the post-cat layer to capture relationships between the different feature types. We found that a 4096 dimensional fully connected layer seems to help increase mean AP slightly, but decreases both normalized accuracy rates due to overfitting. Again, as observed previously, adding additional depth here also causes the model to overfit, and decreases performance.

\begin{table}
\begin{center}
\resizebox{1\linewidth}{!}{
\begin{tabular}{|c|c|c|c|}
\hline
Method & Mean AP & Acc@1 & Acc@5 \\
\hline \hline
Image only & 36.82\% & 39.45\% & 70.15\% \\
\hline
Image + Hashtag context feature & 39.86\% & 42.27\% & 73.38\% \\
Image + Hashtag context feature RL5 & 40.19\% & 42.52\% & 73.57\% \\
Image + Hashtag context feature RL10 & \textbf{40.80\%} & \textbf{43.10\%} & \textbf{74.15\%} \\
\hline \hline
Image + Visual context feature & 38.81\% & 41.53\% & 72.31\% \\
Image + Visual context feature RL5 & 38.75\% & 41.31\% & 72.08\% \\
Image + Visual context feature RL10 & \textbf{39.07\%} & \textbf{41.78\%} & \textbf{72.48\%} \\
\hline \hline
Image + All features with 256/- & 42.34\% & 44.82\% & 75.86\% \\
Image + All features with 256/- RL10 & 42.91\% & \textbf{45.17\%} & \textbf{76.09\%} \\
Image + All features with 256/4096 & 43.28\% & 43.74\% & 74.30\% \\
Image + All features with 256/4096 RL10 & \textbf{43.78\%} & 44.14\% & 74.70\% \\
\hline
\end{tabular}
}
\end{center}
\caption{Results through learning the optimal pooling radius. RL5 and RL10 refer to the number of replicas (5,10) of the radius learning layer used to replace the concatenated histograms. \vspace{-4mm}}
\label{table:radius}
\end{table}

\begin{figure*}
\begin{center}
\includegraphics[width=1\linewidth]{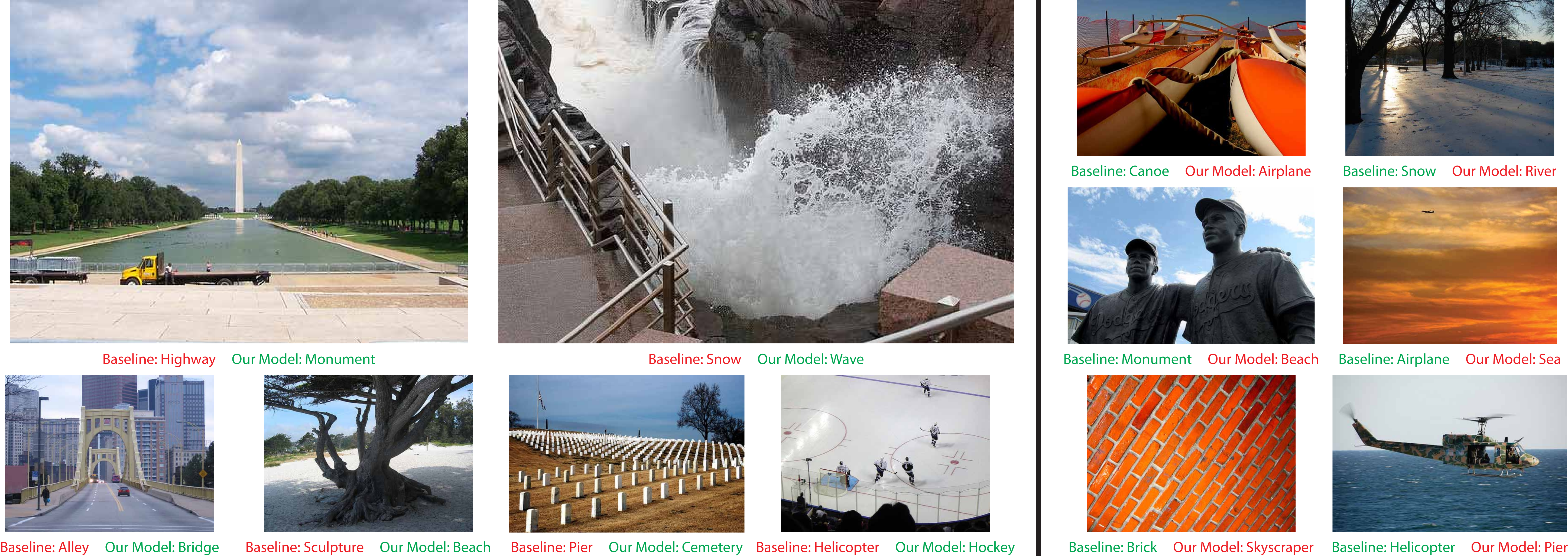}
\end{center}
   \caption{Example results comparing the baseline image model to our best model (256/- RL10), with correct predictions in green and incorrect predictions in red. Image credits given in supplementary material. \vspace{-4mm}}
\label{fig:results}
\end{figure*}

\begin{figure}[t]
\begin{center}
\includegraphics[width=1\linewidth, height=3.5cm]{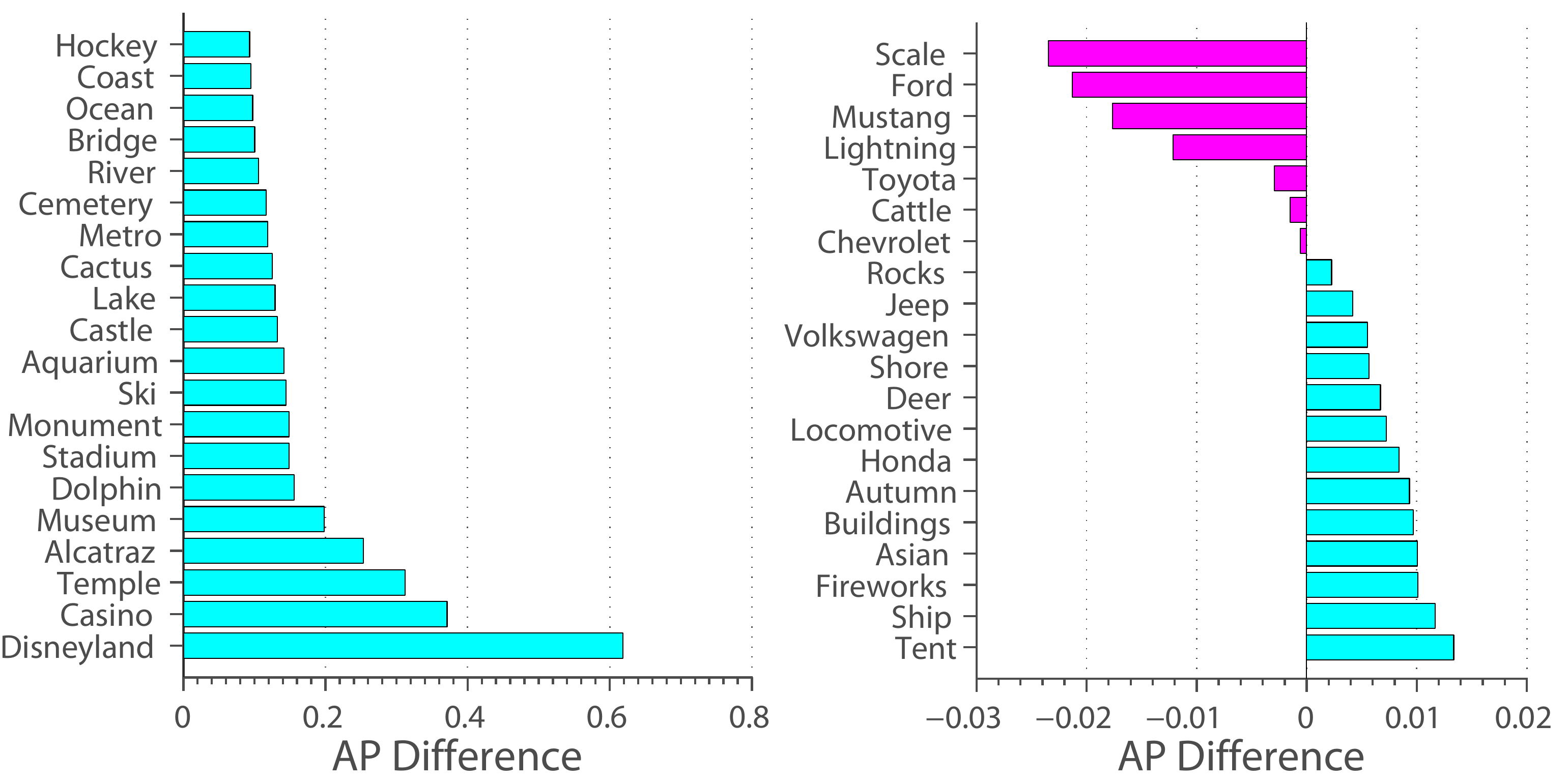}
\end{center}
   \caption{AP difference between our best model (256/- RL10) and the baseline image model for the 20 best and worst classes. \vspace{-1mm}}
\label{fig:apdiff}
\end{figure}

\begin{figure}[t]
\begin{center}
\includegraphics[width=1\linewidth, height=2cm]{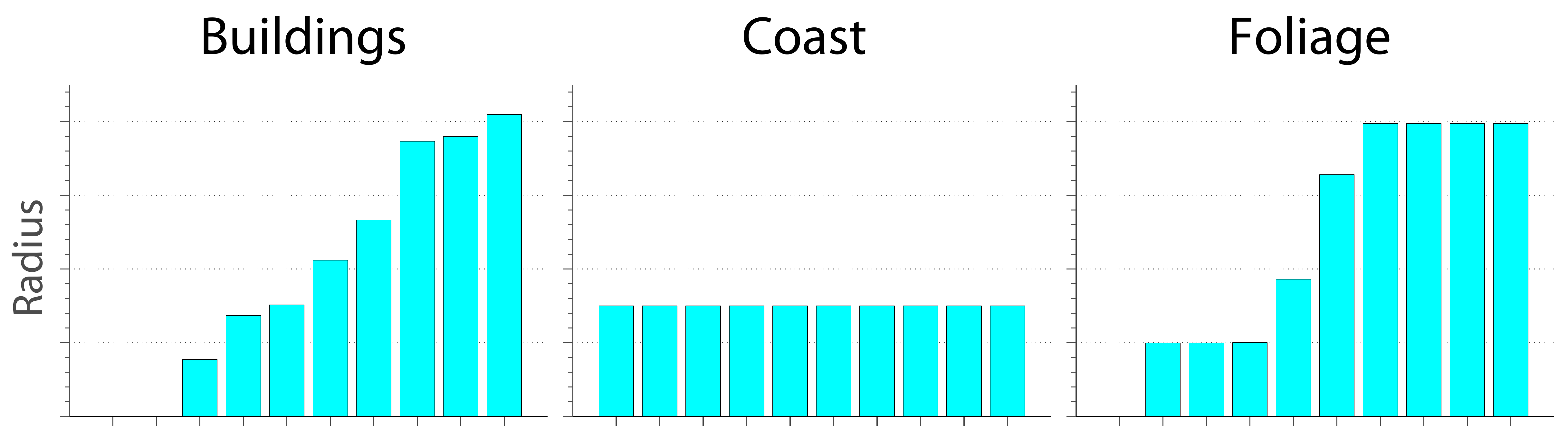}
\end{center}
   \caption{Visualizations of the learned radii for three classes from our best model (256/- RL10), sorted from smallest to largest. \vspace{-4mm}}
\label{fig:radii}
\end{figure}

\subsection{Learning the optimal pooling radius}
In the previous sections, we concatenated histograms computed at varying radii for the hashtag context and visual context features. Since there are often multiple radii and weightings between the radii that are most informative, we replace the concatenated histograms with multiple replicas of radius learning layers, with results shown in Table~\ref{table:radius}. In the top two sections, we observe large improvements for the hashtag context feature, and mild improvements for the visual context feature in a controlled setting with no pre-cat and post-cat layers. In the bottom section, we are able to obtain an additional 0.5\% gain in mean AP by using the radius learning layers for both of our best models from the previous section. We found again that adding the post-cat layer causes the model to slightly overfit, and thus use the 256/- RL10 model as our best model in the following analyses. Figure~\ref{fig:results} shows some interesting examples and predictions.

\vspace{-3mm}

\paragraph{Best and worst classes.}
In Figure~\ref{fig:apdiff}, we show the top 20 best and worst performing classes compared to the baseline image model. Location-specific classes like ``disneyland", ``casino", and ``alcatraz" see a large increase in performance, as they are confined to one or a small number of locations in the United States. On the other hand, some of the worst performing classes are car brands, which suggests that fine-grained car classes are not very location-specific, or not handled well in our features and model. However, since our method for selecting location-sensitive concepts was data-driven and unsupervised, they were included. 

\vspace{-3mm}

\paragraph{Learned radius parameters.}
We visualize the radius parameters learned for several classes in Figure~\ref{fig:radii}. We found that for most concepts, the 10 different replicas of the radius learning layers typically converge to 3 or fewer different radii, like the ``coast" and ``foliage" classes, which suggests that certain radii are indeed more informative. Occasionally, some classes like ``building" learn almost all different radii, possibly because within urban areas the abundance of buildings makes smaller radii important, and within rural areas larger radii become important.

\section{Conclusion}
In this paper, we introduce the problem of image classification with location context. To represent location context, we propose 5 features that help capture context about a particular location, and show how to incorporate them into a CNN model. For features that require pooling over radii, we show how to automatically learn the optimal radius within the same framework, allowing us to obtain better performance and a deeper understanding into the network parameters. Furthermore, we introduce and make publicly available the YFCC100M-GEO100 dataset, which we manually annotate to obtain class labels for geotagged images.

For future work, we would like to explore taking advantage of other aspects of images that are now becoming widely available, such as time and date taken or the social relationships between the users.


\end{document}